\documentclass[letterpaper, 10 pt, conference]{ieeeconf}  

\IEEEoverridecommandlockouts                              

\usepackage{booktabs}
\usepackage{graphicx}
\usepackage[table,xcdraw]{xcolor}
\usepackage{graphics} 
\usepackage{epsfig} 
\usepackage{cite}
\usepackage{times} 
\usepackage{amsmath} 
\usepackage{amssymb}  
\usepackage[switch]{lineno}
\usepackage{mathrsfs}   
\usepackage{bm} 
\usepackage{caption}
\usepackage{subcaption}
\usepackage{url}
\usepackage{color}
\usepackage{algorithm}
\usepackage{algorithmic}

\usepackage{amsthm}
\usepackage{svg}

\usepackage{wrapfig}
\let\labelindent\relax
\usepackage{enumitem}
\usepackage{makecell} 
\usepackage{mathtools}
\usepackage{hyperref}
\usepackage{ifthen}
\usepackage{xcolor, soul}
\sethlcolor{yellow}
\DeclareRobustCommand{\hlcyan}[1]{{\sethlcolor{cyan}\hl{#1}}}

\newcommand{\highlightedequation}[1]{%
        \colorbox{cyan}{%
            \begin{minipage}{\dimexpr\linewidth-2\fboxsep\relax}%
                #1%
            \end{minipage}%
        }%
}

\newcounter{matriz}

\newcounter{tableeqn}[table]

\newcounter{tablesubeqn}[tableeqn]

\usepackage{etoolbox}
\AtBeginEnvironment{align}{\setcounter{subeqn}{0}}
\newcounter{subeqn} %

\usepackage{adjustbox}

\newtheorem{problem}{Problem}

\newtheorem{remark}{Remark}

\newtheorem*{assumption*}{Assumption}
\newtheorem*{opt*}{Policy Repair Optimization}
\newtheorem*{opt1*}{Safety-guided Policy Repair Optimization}
\newtheorem*{opt2*}{SARP Optimization}

\newcommand{\calP}{\mathcal{Z}}
\newcommand{\calS}{\mathcal{S}}
\newcommand{\calA}{\mathcal{A}}
\newcommand{\NcalP}{n_{\mathcal{Z}}}
\newcommand{\NcalS}{n_{\mathcal{S}}}
\newcommand{\NcalA}{n_{\mathcal{A}}}
\newcommand{\feat}{z}
\title{\LARGE \bf
Repairing Neural Networks for Safety in Robotic Systems using Predictive Models
}

\author{Keyvan Majd$^{1}$, Geoffrey Clark$^{1}$, Georgios Fainekos$^{2}$, and Heni Ben Amor$^{1}$ 
\thanks{$^{1}$ K. Majd, G. Clark, and H. Ben Amor 
        {\tt\small \{majd, gmclark1, hbenamor\}@asu.edu} are with SCAI, Arizona State University, Tempe, AZ, USA.}%
\thanks{$^{2}$ G. Fainekos is with Toyota NA R\&D, Ann Arbor, MI, USA.}
}

\newboolean{highlight}
\setboolean{highlight}{false} 
\newboolean{extend}
\setboolean{extend}{false} 

\begin{document}

\maketitle

\begin{abstract}

This paper introduces a new method for safety-aware robot learning, focusing on repairing policies using predictive models. 
Our method combines behavioral cloning with neural network repair in a two-step supervised learning framework. 
It first learns a policy from expert demonstrations and then applies repair subject to predictive models to enforce safety constraints.
The predictive models can encompass various aspects relevant to robot learning applications, such as proprioceptive states and collision likelihood. 
Our experimental results demonstrate that the learned policy successfully adheres to a predefined set of safety constraints on two applications: mobile robot navigation, and real-world lower-leg prostheses.
Additionally, we have shown that our method effectively reduces repeated interaction with the robot, leading to substantial time savings during the learning process.

\end{abstract}


\section{Introduction}
\label{sec:introduction}
\ifthenelse{\boolean{highlight}}{
\hlcyan{
Robot learning holds great promise for advancing wearable robotics, such as Prosthetics, Orthoses, and Exoskeletons~\mbox{\cite{kyrarini2021survey}}. 
These devices are typically designed with a one-size-fits-all approach. 
Deep learning techniques offer the potential to customize policies for each user that leads to improved comfort and quality of life.
One common approach for customizing policies has been through reinforcement learning (RL). 
However, RL can be time-consuming and potentially hazardous in real-world scenarios, as agents must explore the environment and learn from mistakes ~\mbox{\cite{brunke2022safe}}. 
In our previous work~\mbox{\cite{majd2022safe}}, we introduced an algorithm for training neural network policies that satisfy given safety specifications using neural network repair. 
By solving a layer-wise Mixed-integer Quadratic Program (MIQP), we modified network weights to meet desired safety constraints on actions while providing theoretical guarantees of safety for the samples used in repair.
However, our method primarily focused on addressing explicit safety constraints over the output space of the neural network.
Building upon our previous work, we now extend our method to address implicit safety constraints using predictive models. 
}
}{
Robot learning holds great promise for advancing wearable robotics, such as Prosthetics, Orthoses, and Exoskeletons \cite{kyrarini2021survey}. 
These devices are typically designed with a one-size-fits-all approach. 
Deep learning techniques offer the potential to customize policies for each user that leads to improved comfort and quality of life.
One common approach for customizing policies has been through reinforcement learning (RL). 
However, RL can be time-consuming and potentially hazardous in real-world scenarios, as agents must explore the environment and learn from mistakes \cite{brunke2022safe}.
In our previous work \cite{majd2022safe}, we introduced an algorithm for training neural network policies that satisfy given safety specifications using neural network repair. 
By solving a layer-wise Mixed-integer Quadratic Program (MIQP), we modified network weights to meet desired safety constraints on outputs while providing mathematical guarantees on safety for the samples used in repair.
However, our method primarily focused on addressing explicit safety constraints over the output space of the neural network.
Building upon our previous work, we now extend our method to address implicit safety constraints on the output of predictive models. 
}

\begin{figure*}
    \vspace{7pt}
    \centering
    \includegraphics[width=0.86\textwidth]{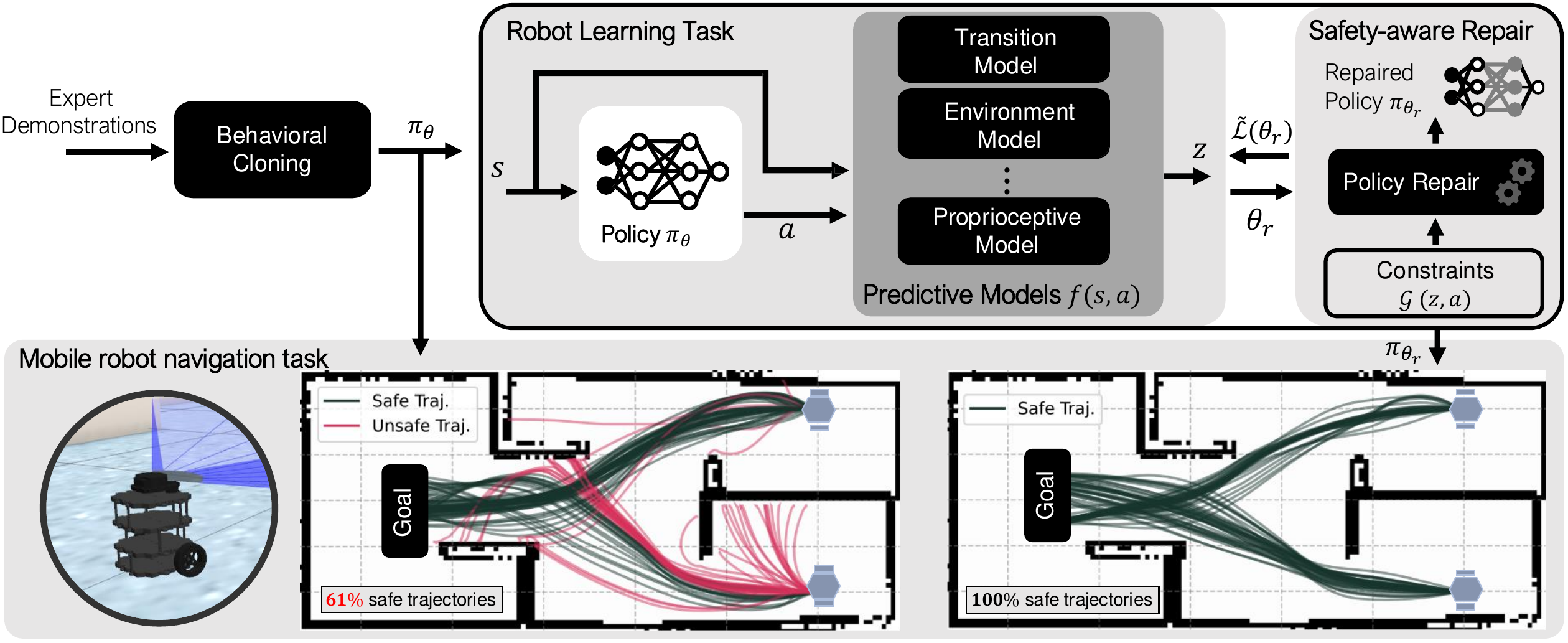}
    \caption{Safety-Aware Repair with Predictive models (SARP). Left: A policy trained for a mobile navigation task (point to goal motion planning in this figure). Right: Policy repair module that adjusts the policy parameters to penalize unsafe behavior, based on a set of safety constraints and the loss of predictive models. Predictive models may include state-action transition model, a model of environment, or a proprioceptive model predicting internal states of the system.}
    \label{fig:teaser}
    \vspace{-7pt}
\end{figure*}

In this paper, we present \textbf{SARP} (\textbf{S}afety-\textbf{A}ware \textbf{R}epair with \textbf{P}redictive models), 
an approach for safety-driven learning of robot policies from human demonstrations. 
Our method focuses on repairing an existing neural network policy to ensure compliance with a defined set of safety constraints. 
We leverage a predictive model to predict the features of the system given states and actions. 
These features can encompass various aspects in different robot learning applications, such as proprioceptive states in biomechanical applications, or collision occurrence in navigation scenarios.

\ifthenelse{\boolean{highlight}}{
\hlcyan{In a two-step supervised learning process, SARP initially trains a policy using expert demonstrations. 
We then employ a predictive model to enforce safety constraints on the policy. 
This is achieved through the application of neural network repair techniques to regulate the predicted system features. 
SARP addresses both implicit constraints on the policy, using differentiable predictive models, and explicit bounding constraints on actions.
We assess our method in two safety-critical case studies. 
We first compare SARP's performance with two state-of-the-art safe RL methods, as presented in \!\mbox{\cite{achiam2017constrained}} and \!\mbox{\cite{dalal2018safe}}. 
This comparison aims to demonstrate SARP's efficacy and comparable performance in a simulated mobile robot navigation task as a showcase example.
Our goal is not to position SARP as a substitute for RL but rather as a complementary method that addresses the safety concerns associated with lengthy simulation in RL for safety-critical robot learning tasks, such as lower-leg prosthesis control.
We finally test SARP for controlling a real-world lower-leg prosthesis in an IRB-approved study.
}
}
{
In a two-step supervised learning process, SARP initially trains a policy using expert demonstrations. 
We then employ a predictive model to enforce safety constraints on the policy. 
This is achieved through the application of neural network repair techniques to regulate the predicted system features. 
SARP addresses both implicit constraints on the policy, using differentiable predictive models, and explicit bounding constraints on actions.
We assess our method in two safety-critical case studies. 
We first compare SARP's performance with two state-of-the-art safe RL methods, as presented in \cite{achiam2017constrained} and \cite{dalal2018safe}. 
This comparison aims to demonstrate SARP's efficacy and comparable performance in a simulated mobile robot navigation task as a showcase example.
Our goal is not to position SARP as a substitute for RL but rather as a complementary method that addresses the safety concerns associated with lengthy simulation in RL for safety-critical robot learning tasks, such as lower-leg prosthesis control.
We finally test SARP for controlling a real-world lower-leg prosthesis in an IRB-approved study.
}

Our novel \textbf{contribution} lies in repairing the policy to ensure that the predicted features of the system adhere to a predefined set of safety constraints. 
Specifically, this paper makes the following contributions:

\begin{enumerate}
    \item  We propose a novel two-step supervised learning method that combines imitation learning and neural network repair using predictive models.
    \item Our framework can utilize any pre-existing differentiable model in the repair process that reduces the repetitive interactions with the robot.
    \item We compare SARP with two safe RL-based methods of constrained policy optimization (CPO) in \cite{achiam2017constrained} and the safety layer approach in \cite{dalal2018safe}.
    \item The code for reproducing the results is available at \url{https://github.com/k1majd/SARP.git}.
\end{enumerate}

\section{Related Work}
\subsection{Safe Reinforcement Learning}

Safety in robot learning has been extensively explored in the existing literature \cite{brunke2022safe}.
One category of methods focuses on learning uncertain dynamics certifying safety. 
\ifthenelse{\boolean{extend}}
{These methods typically rely on prior knowledge of the system dynamics \cite{liu2022robot, alshiekh2018safe} or learn the uncertainties in the dynamics \cite{choi2020reinforcement, chen2021context, as2021constrained}.
Mostly, these techniques provide certification of safety by ensuring the safety set invariance and reachability 
\cite{robey2020learning, taylor2020learning, chang2019neural},
 or through temporal logic \cite{aksaray2016q, hasanbeig2019reinforcement, sadigh2014learning}.}
{\ifthenelse{\boolean{highlight}}{\hl{These methods typically rely on prior knowledge of the system dynamics\mbox{\cite{liu2022robot, alshiekh2018safe,dawson2023safe}} or learn the uncertainties in the dynamics\mbox{\cite{choi2020reinforcement, chen2021context, as2021constrained}}.}}{These methods typically rely on prior knowledge of the system dynamics \cite{liu2022robot, alshiekh2018safe,dawson2023safe} or learn the uncertainties in the dynamics \cite{choi2020reinforcement, chen2021context, as2021constrained}.}}
\ifthenelse{\boolean{extend}}
{Another category of methods focuses on ensuring safety without prior knowledge of robot dynamics or safety constraints.
These approaches either encourage safety during policy exploration or penalize dangerous actions.}
{\ifthenelse{\boolean{highlight}}
    {\hl{Another category encourages safety during exploration or penalizes risky actions in a model-free fashion.}}
    {Another category encourages safety during policy exploration or penalizes dangerous actions in a model-free fashion.}}
\ifthenelse{\boolean{extend}}
{One safety exploration technique is to learn a safety critic function to detect potentially unsafe actions \cite{thananjeyan2021recovery}. 
Some methods incorporate uncertainty or risk into the learning process. 
For example, \cite{chua2018deep} uses an ensemble model to estimate uncertainty and enable risk-averse actions, 
while \cite{thananjeyan2020safety} trains an ensemble model with expert demonstrations and applies Model Predictive Control (MPC) for safe exploration. 
Methods in \cite{dalal2018safe} and \cite{yu2022towards} also perform safe exploration through learning a safety layer or a safety editor policy, respectively.}
{\ifthenelse{\boolean{highlight}}
    {\hl{Safety exploration strategies include learning a safety critic function\mbox{\cite{thananjeyan2021recovery}}, taking risk-averse action through learning an ensemble model\mbox{\cite{chua2018deep, thananjeyan2020safety}}, or editing the policy through a safety layer\mbox{\cite{dalal2018safe, yu2022towards}}.}}
    {Safety exploration strategies include learning a safety critic function \cite{thananjeyan2021recovery}, taking risk-averse action through learning an ensemble model\cite{chua2018deep, thananjeyan2020safety}, or editing the policy through a safety layer \cite{dalal2018safe, yu2022towards}.}}
\ifthenelse{\boolean{extend}}
{A closely related line of work to ours involves employing RL approaches to solve a relaxed Constrained Markov Decision Process (CMDP) using Lagrangian relaxation \cite{wright2006numerical}.
Constrained Policy Optimization (CPO) \cite{achiam2017constrained}, Interior-point Policy Optimization (IPO) \cite{liu2020ipo}, and Primal-Dual Policy Optimization (PDO) \cite{chow2017risk} are widely used policy optimization techniques that address this relaxed problem. 
These techniques perform primal-dual updates over derived surrogates of the reward and constraint cost functions, 
guaranteeing monotonic policy improvement and near-constraint satisfaction at each iteration.
In a different policy update approach, \cite{chow2019lyapunov} employs Lyapunov functions to project continuous actions, ensuring a decrease in the Lyapunov function after each update step.}
{\ifthenelse{\boolean{highlight}}
    {\hl{A closely related line of work to ours involves employing RL approaches to solve a relaxed Constrained Markov Decision Process (CMDP) using Lagrangian relaxation\mbox{\cite{achiam2017constrained, liu2020ipo, chow2017risk}}. These techniques use primal-dual updates on reward and constraint surrogates to ensure policy improvement and near-constraint satisfaction.}}
    {A closely related line of work to ours involves employing RL approaches to solve a relaxed Constrained Markov Decision Process (CMDP) using Lagrangian relaxation \cite{achiam2017constrained, liu2020ipo, chow2017risk}.
    These techniques use primal-dual updates on reward and constraint surrogates to ensure policy improvement and near-constraint satisfaction.}}

\ifthenelse{\boolean{extend}}
{In RL, finding an effective reward-shaping scheme that promotes safe behavior while achieving desired performance objectives is a non-trivial task. 
Actions that lead to high rewards may also increase the risk of safety violations. 
Moreover, RL methods require extensive simulation or robot-environment interaction time to converge and to learn optimal policies. 
Such time-consuming learning process is especially problematic in safety-critical scenarios involving physical human-robot interaction, such as with robotic prostheses devices. 
Repeated interactions between humans and robots to reach a safe solution can pose a risk to the human user. 
In contrast, SARP combines Behavioral Cloning (BC) with neural network repair in a two-step supervised learning framework. 
Our method reduces the necessity for lengthy simulations, allowing for easier offline repair of policy for safety. 
SARP adds to the tools of robotics engineers when RL approaches are deemed too expensive or potentially unsafe, such as in physical human-robot interactions.}
{\ifthenelse{\boolean{highlight}}
{\hlcyan{In RL, finding an effective reward-shaping scheme that promotes safe behavior while achieving desired performance objectives is a non-trivial task. 
Actions that lead to high rewards may also increase the risk of safety violations. 
Moreover, RL methods require extensive simulation or robot-environment interaction time to converge and learn optimal policies. 
This time-consuming learning process is especially problematic in safety-critical scenarios involving human-robot interaction, such as with prostheses. 
Repeated interactions between humans and robots to reach a safe solution can pose a risk to the human user. 
In contrast, SARP combines Behavioral Cloning (BC) with neural network repair in a two-step supervised learning framework. 
Our method reduces the necessity for lengthy simulations, allowing for easier offline policy evaluation and ensuring safety. 
Therefore, SARP provides a reliable and practical solution for ongoing policy monitoring and repair. 
Our objective is not to present SARP as a substitute for RL but rather as a safe method that can achieve comparable performance to RL in applications where extensive exploration of safety policies is not feasible, such as in prosthesis control.
}}
{
In RL, finding an effective reward-shaping scheme that promotes safe behavior while achieving desired performance objectives is a non-trivial task. 
Actions that lead to high rewards may also increase the risk of safety violations. 
Moreover, RL methods require extensive simulation or robot-environment interaction time to converge and to learn optimal policies. 
Such time-consuming learning process is especially problematic in safety-critical scenarios involving physical human-robot interaction, such as with robotic prostheses devices. 
Repeated interactions between humans and robots to reach a safe solution can pose a risk to the human user. 
In contrast, SARP combines Behavioral Cloning (BC) with neural network repair in a two-step supervised learning framework. 
Our method reduces the necessity for lengthy simulations, allowing for easier offline repair of policy for safety. 
SARP adds to the tools of robotics engineers when RL approaches are deemed too expensive or potentially unsafe, such as in physical human-robot interactions.}}

\ifthenelse{\boolean{highlight}}{
\subsection{\colorbox{yellow} {Neural Network Repair}}
}{
\subsection{Neural Network Repair}
}
\ifthenelse{\boolean{extend}}
{Another relevant area of research related to this paper is neural network (NN) repair. 
This involves collecting counter-examples through verifying and testing NNs \cite{TjengXT2019iclr, LiuEtAl2021fto}, 
and making minimal weight modifications to ensure compliance with specific safety constraints for erroneous behaviors.
One approach to repair is repeated retraining and fine-tuning based on counterexamples \cite{sinitsin2020editable,ren2020few, DongEtAl2021qrs}. 
However, this approach requires the availability of counterexample labels, which may involve additional data collection. 
Several approaches exist for provable repair of networks that can guarantee safety for either faulty samples or faulty input-output linear regions of the network \cite{goldberger2020minimal,FuLi2022iclr,yang2021neural, sotoudeh2021provable, majd2022safe, majd2022certifiably, DongEtAl2021qrs}. 
These methods are only applicable to networks with piecewise linear activation functions and only address explicit constraints over the network's output. 

Our repair method extends beyond the scope of piecewise-linear networks and offers repair to both explicit and implicit constraints on the output of the network.
We propose employing a differentiable predictive model in repairing a pre-trained NN policy. 
It enables the incorporation of task-specific features such as collision prediction or proprioceptive features as implicit constraints on the policy. }
{Another relevant area of research related to this paper is neural network (NN) repair. 
One approach to repair is repeated retraining and fine-tuning based on counterexamples \cite{sinitsin2020editable,ren2020few, DongEtAl2021qrs}. 
However, this approach requires the availability of counterexample labels, which may involve additional data collection. 
Several approaches exist for provable repair of networks that can guarantee safety for either faulty samples or faulty input-output linear regions of the network \cite{goldberger2020minimal,FuLi2022iclr,yang2021neural, sotoudeh2021provable, majd2022safe, majd2022certifiably, DongEtAl2021qrs}. 
These methods are only applicable to networks with piecewise linear activation functions and only address explicit constraints over the network's output. 

Our repair method extends beyond the scope of piecewise-linear networks and offers repair to both explicit and implicit constraints on the output of the network.
We propose employing a differentiable predictive model in repairing a pre-trained NN policy. 
It enables the incorporation of task-specific features such as collision prediction or proprioceptive features as implicit constraints on the policy. }

\section{Problem Formulation}

While learning the policy directly from expert demonstrations may not guarantee a safe policy, safety can be attained by specifying and enforcing safety constraints that the policy must adhere to. 
Our method guides the policy toward safety through predictive models that predict the future system features. 
A predictive model can have different forms.
It might be derived analytically or learned from expert demonstrations and robot exploration samples.
A predictive model can learn static environmental properties such as  friction and surface characteristics, or it may be a proprioceptive model that estimates the internal states based on sensor outputs.

Our approach in this work proposes a safety-guided policy repair method that employs a set of safety constraints on either the predictive features of system or actions of the policy. 
Figure~\ref{fig:teaser} shows an example scenario in which a robot learns from demonstrations to navigate from multiple start locations to a goal. 
First, we pre-train a NN policy $\pi_{\theta}$ with parameters $\theta\in\Theta$ using imitation learning to learn from expert demonstrations, where $\Theta$ represents the network's parameter space.  
As shown in Fig. \ref{fig:teaser}, the pre-trained policy $\pi_{\theta}$  generates about $39\%$ unsafe trajectories, i.e., robot actions lead to collisions. 
Next, we employ a differentiable predictive model that predicts the occurrence of collision in future time steps to guide the pre-trained policy towards safety. This is achieved by optimizing the policy parameters $\theta$ through a NN repair module that penalizes unsafe behavior, as shown in Fig. \ref{fig:teaser} (right). 
This safety-guided repair step is performed in an offline optimization and does not involve iterative interaction with the environment. 
The resulting repaired parameters $\theta_r$ lead to a policy that satisfies the safety constraints over the training data.
\begin{problem}[Safety-Aware Repair with Predictive models] \label{prob1}
Let
$f\!\!:\calS\times \calA\rightarrow \calP$ be a differentiable predictive model that predicts a feature $\feat \in \calP\subseteq \mathbb{R}^{\NcalP}$ of the system given the previous state $s\in\calS\subseteq \mathbb{R}^{\NcalS}$ and action $a\in\calA\subseteq \mathbb{R}^{\NcalA}$. 
Here, $\calP$, $\calS$, and $\calA$ denote the feature, state, and action spaces, respectively, with dimensions $\NcalP$, $\NcalS$, and $\NcalA$.
Consider a trained NN policy $\pi_{\theta}:\calS\rightarrow\calA$, and a differentiable predictive model $f(s, a)$, 
which predicts a system feature $\feat$. 
Let $\mathcal{G}(\feat,a)$ denote a set of constraints that must hold true for the robot learning task. 
 The goal of safety-aware repair is to adjust the policy parameters $\theta$ so as to minimize the loss function $\mathcal{L}: \calS\times\calA \rightarrow \mathbb{R}$ while satisfying $\mathcal{G}(\feat,a)$.
\end{problem}

\section{ Safety-guided Policy Repair using Predictive Models} \label{sec: safet-guided repair}
\ifthenelse{\boolean{extend}}{This section presents the formulation of SARP.}{}
Our method involves formulating and solving an optimization problem that minimizes a loss function subject to safety constraints over the predicted system features and actions. 
The features of the system could represent a non-observable or future state of the system, or other system parameters, such as friction, mass, etc.
SARP aims to modify the parameters $\theta$ of the policy $\pi_{\theta}$ to adjust the feature
$\feat = f(s,a)$ and the actions $a=\pi_{\theta}(s)$ such that the set of constraints $\mathcal{G}(f(s,a),a)$ is satisfied for all $(s,a)\in\calS\times\calA$. 

Due to the non-convex nature of the optimization problem and the nonlinear activation functions of the policy, finding a global solution can be challenging.
To address this issue, the works in~\cite{majd2022safe} and \cite{majd2022certifiably} proposed a layer-wise relaxation approach that focuses on repairing the networks with Rectified Linear Unit (ReLU) activation functions. 
In particular, assuming quadratic loss functions and modifying the weights of a single layer, 
this method formulates repair as a mixed-integer quadratic program (MIQP) that can be solved using off-the-shelf solvers \cite{majd2022safe,majd2022certifiably}.
This method also assumes explicit constraints on the output of policy network.

This paper presents an extended approach to repairing the policy network $\pi_{\theta}$, inspired by the previous methodologies \cite{majd2022safe} and \cite{majd2022certifiably}. 
In addition to addressing explicit constraints on the output of the policy network, our method includes constraints associated with the system's features $\feat$, as defined by the predictive model $f(s,a)$. 
Unlike \cite{majd2022safe} and \cite{majd2022certifiably}, 
which focused on a single layer repair and were limited to only (Piecewise-)Linear activation functions, our method offers a more comprehensive solution. 
We do not restrict ourselves to a particular class of activation functions and we enable repair across multiple layers of the policy network.
We can formulate Problem \ref{prob1} as an optimization problem:
\ifthenelse{\boolean{extend}}
{\begin{align} 
\theta_r = \underset{\theta\in\Theta}{\arg\min}&~\mathcal{L}(\pi_\theta) \label{eq: cost}\\ 
\mbox{ s.t.} &~\feat = f(s,\pi_{\theta})~~\forall s\in\calS, \label{eq: const. theta_s}\\
&~\mathcal{G}(\feat, \pi_{\theta}). \label{eq: predicate}
\end{align}}
{
\begin{align} 
\theta_r = \underset{\theta\in\Theta}{\arg\min}&~\mathcal{L}(\pi_\theta) \label{eq: cost}\\ 
\mbox{ s.t.} &~\feat = f(s,\pi_{\theta})~~\forall s\in\calS, \label{eq: const. theta_s}\\
&~\mathcal{G}(\feat, \pi_{\theta}). \label{eq: predicate}
\end{align}
}

\begin{figure}
\vspace{-2.1mm}
\end{figure}
\begin{algorithm}[tb]
\caption{Safety-aware repair with predictive models}
\label{alg:verifier2}
\begin{algorithmic}[1]
\renewcommand{\algorithmicrequire}{\textbf{Input:}}
\renewcommand{\algorithmicensure}{\textbf{Output:}}
\REQUIRE $\pi_{\theta}, f, \Tilde{\mathcal{L}}, \mathcal{G}, \mathcal{D}$
\ENSURE  $\pi_{\theta_r}$
\STATE Initialize $\mu_0 \in \mathbb{R}_+, \lambda_0 \leftarrow 0, k\leftarrow0, \theta_r\leftarrow \theta$
\WHILE{$\textsc{SafetyCheck}(\theta_r, f, \mathcal{G}, \mathcal{D})\neq$ \TRUE}
\STATE $\theta_r \leftarrow \arg \min_{\theta_r} \Tilde{\mathcal{L}}(\pi_{\theta}(s), \lambda_k, \mu_k)$, for $(s, a) \in \mathcal{D}$ 
\STATE $\lambda_{k+1}\leftarrow  -ReLU\Big(\eta\sum^C_{c=1}g_c(f(s,\pi_{\theta_r}(s))) - \lambda_k\Big)$ \label{alg: lag update}
\STATE $\mu_{k+1} \leftarrow \beta\mu_k$ \label{alg: penalty update}
\STATE $k\leftarrow k+1$
\ENDWHILE
\end{algorithmic}
\end{algorithm}
This problem optimizes $\theta$ by minimizing the loss function (\ref{eq: cost}), while satisfying the constraints (\ref{eq: const. theta_s}) and (\ref{eq: predicate}).
In this paper, we assume that the set of constraints (\ref{eq: predicate}) is defined as a conjunction of equality constraints of form $\bigwedge_{c=1}^C{g_c(\feat,\pi_{\theta})= 0}$, where $C$ represents the number of constraints and $g_c:\calP\times\calA\rightarrow \mathbb{R}$ is a differentiable nonlinear function dependent on variables $\feat$ and $a$.
Since constraints in practical scenarios often involve inequalities, we handle inequality constraints of the form $g(\cdot) \leq 0$, we employ a replacement approach by transforming them into a rectified linear unit (ReLU) function: $ReLU(g(\cdot)) = \max\{0, g(\cdot)\}$. 
This modification effectively allows us to address the inequality constraints, as $ReLU(g(\cdot)) = 0 \Longleftrightarrow g(\cdot) \leq 0$. 
Given that $\feat=f(s,\pi_{\theta}(s))$, we can integrate the constraints (\ref{eq: const. theta_s}) and (\ref{eq: predicate}) by $\bigwedge_{c=1}^C{g_c(f(s,\pi_{\theta}(s)),\pi_{\theta}(s))= 0}$.
We further relax the constraints by incorporating them into the loss function (\ref{eq: cost}) inspired by the Augmented Lagrangian Method (ALM) \cite{wright2006numerical}. 
The modified loss with incorporated constraints is given by:
\ifthenelse{\boolean{highlight}}{\highlightedequation{
\begin{align}
\Tilde{\mathcal{L}}(s,\lambda, \mu) =  \mathcal{L}(s, \pi_\theta(s))&-\sum^C_{c=1}\lambda_c g_c\Big(f(s,\pi_{\theta}(s))\Big)\nonumber\\
&+\frac{\mu}{2}\sum^C_{c=1}g^2_c\Big(f(s,\pi_{\theta}(s))\Big).\label{eq: augmented loss}
\end{align}}
}{\begin{align}
\Tilde{\mathcal{L}}(\theta,\lambda, \mu) =  \mathcal{L}(\pi_\theta)&-\sum^C_{c=1}\lambda_c g_c\Big(f(s,\pi_{\theta}),\pi_{\theta}\Big)\nonumber\\
&+\frac{\mu}{2}\sum^C_{c=1}g^2_c\Big(f(s,\pi_{\theta}), \pi_{\theta}\Big).\label{eq: augmented loss}
\end{align}}
Here, $\sum^C_{c=1}g_c(\cdot)$ represents the penalty terms with Lagrange multipliers $\lambda\in \mathbb{R}^C$, 
that balance the satisfaction of the constraints and the overall objective optimization. 
The parameter $\mu\in \mathbb{R}$ scales the quadratic penalty terms $\sum^C_{c=1}g^2_c(\cdot)$, 
that penalize the violations of constraints and encourage the optimization process to satisfy them.
\begin{figure}[t]
    \vspace{7pt}
    \centering
    \includegraphics[width=\linewidth]{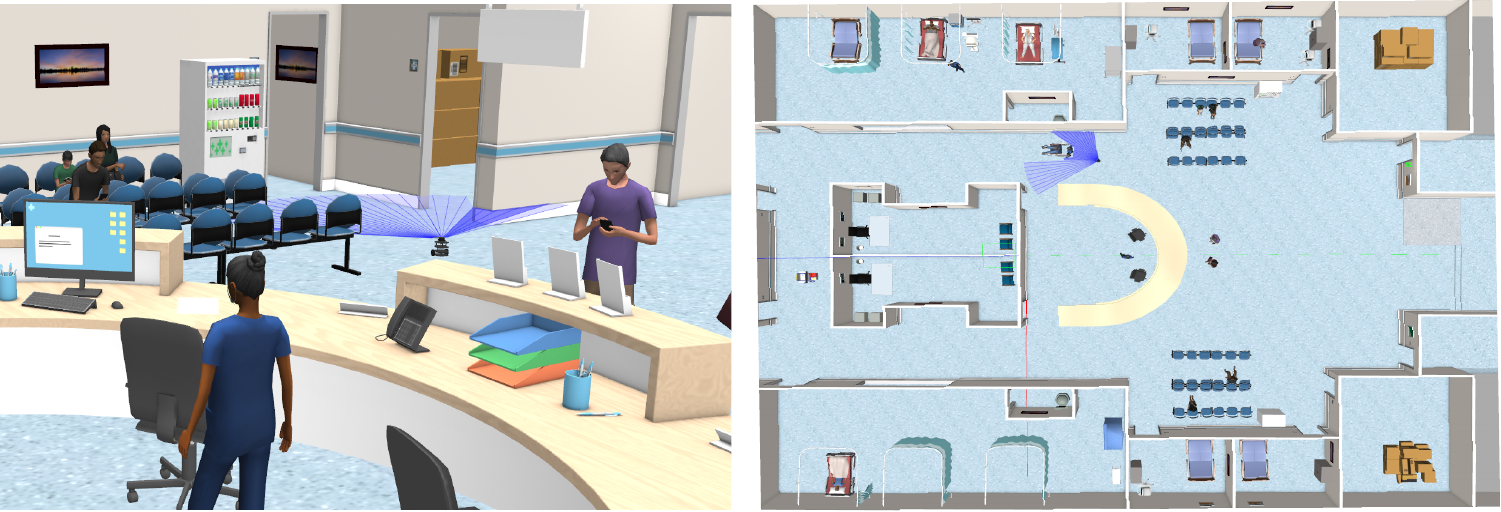}
    \caption{A simulation of a mobile robot in a hospital scenario. The robot is tasked with getting to different rooms without colliding with the environment.}
    \label{fig:hospital}
    \vspace{-7pt}
\end{figure}
We formulate the safety-aware policy repair as follows:
\begin{opt2*}
Let $\pi_{\theta}$ denote a trained policy, and let $\feat$ be a feature of the system given by $f(s, a)$ as defined in (\ref{eq: const. theta_s}).
Let also $\bigwedge_{c=1}^C{g_c(\feat,a)= 0}$ be the 
constraints on the feature $\feat$ for the policy action $a=\pi_{\theta}(s)$, 
where $C$ represents the number of constraints, $g_c(\cdot)$ is a differentiable nonlinear function, and $s\in\calS$.
Safety-aware policy repair optimizes $\theta$ such that the augmented loss function (\ref{eq: augmented loss}) is minimized:
\begin{align} 
& \theta_r=\underset{\theta\in\Theta,\mu\in\mathbb{R},\lambda\in\mathbb{R}^C}{\arg\min} \Tilde{\mathcal{L}}(\theta,\lambda, \mu). \label{eq: cost2}
\end{align}
\end{opt2*}
To solve the problem (\ref{eq: cost2}), we present Alg. \ref{alg:verifier2} 
that is inspired by the iterative primal-dual algorithm \cite{wright2006numerical}. 
Consider a collection of $N$ samples denoted as $\mathcal{D}=\{ (s_n,a_n)\}^N_{n=1}$, that represent the expert's demonstrations. 
Here, $(s_n,a_n)\in \calS\times\calA$, where $\calS$ and $\calA$ denote the sets of states and actions, respectively.
We further define $\pi_{\theta}$ be a trained policy over the expert's samples $\mathcal{D}$. 
In Alg. \ref{alg:verifier2}, 
we initialize $\lambda_0$ as $0$ and $\mu_0$ as a positive real number ($\mu\in\mathbb{R}_+$).
At each iteration $k$, the algorithm modifies the parameters of $\pi_{\theta}$ to minimize the augmented loss (\ref{eq: augmented loss}) for all state-action tuples in $\mathcal{D}$. 
We employ Stochastic Gradient Descent \cite{kingma2014adam} as our chosen optimization algorithm to solve (\ref{eq: cost2}). 
Next, the algorithm updates the Lagrange multiplier $\lambda_k$ by dual ascent as in line \ref{alg: lag update}. 
Additionally, the penalty coefficient $\mu_k$ is updated according to line \ref{alg: penalty update} of the algorithm.
Here, $\eta \in \mathbb{R}_+$ and $\beta\in [1,\infty)$.
\ifthenelse{\boolean{highlight}}{
\hl{To assess the safety of the repaired policy $\pi_{\theta_r}$ on the training set $\mathcal{D}$, 
Alg. {\ref{alg:verifier2}} employs the function $\textsc{SafetyCheck}(\cdot)$ at each iteration. 
This function terminates repair if the safety condition is satisfied for all samples.}
}{
To assess the safety of the repaired policy $\pi_{\theta_r}$ on the training set $\mathcal{D}$, 
Alg. {\ref{alg:verifier2}} employs the function $\textsc{SafetyCheck}(\cdot)$ at each iteration.
This function terminates repair if the safety condition is satisfied for all samples.
}
\ifthenelse{\boolean{highlight}}
{\hl{Note that the primary factor affecting computational cost is the number of optimization variables $\theta$. 
The convergence rate can also be influenced by the number of violated constraints, but the impact is problem-dependent.}}
{Note that the primary factor affecting computational cost is the number of optimization variables $\theta$. 
The convergence rate can also be influenced by the number of violated constraints, but the impact is problem-dependent.}

\begin{remark}
In addition to verifying safety for all samples in $\mathcal{D}$ using $\textsc{SafetyCheck}(\cdot)$, we can further validate that the property $\mathcal{G}$ holds over a broader operating domain $S' \times A' \subseteq S \times A$ using a neural network reachability tool \cite{yang2021neural}. This approach not only confirms constraint satisfaction for regions beyond the collected sample set but also allows us to add any violating samples to $\mathcal{D}$ for further robustness.
\end{remark}

\begin{figure}[t]
\vspace{7pt}
    \centering
    \includegraphics[width=0.86\linewidth]{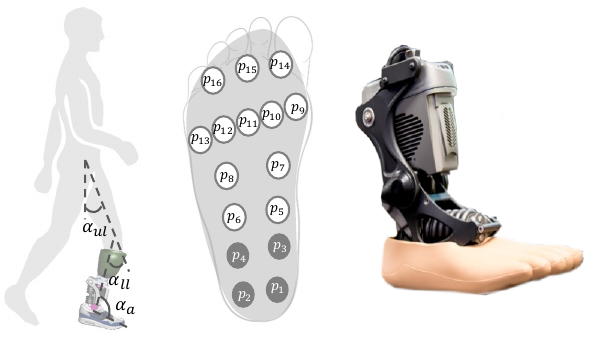}
    \caption{Lower-leg prosthesis system. (left) Image depicts the upper limb angle $\alpha_{ul}$, the lower limb angle $\alpha_{ll}$, and the ankle angle $\alpha_{a}$. (middle) The location of pressure sensors $p_1$-$p_{16}$. (right) The robotic lower-limb prosthesis device.}
    \label{fig:sensor model}
    \vspace{-7pt}
\end{figure}
\section{Application Scenarios}
In this section, we evaluate SARP by presenting its application in two safety-critical scenarios. 
\ifthenelse{\boolean{highlight}}
{\hlcyan{
The first example is presented as a simulation showcase, comparing SARP with two state-of-the-art safe RL approaches: Constrained Policy Optimization (CPO) \mbox{\cite{achiam2017constrained}} and the safety-layer approach (DDPG+SL) \mbox{\cite{dalal2018safe}}.
In this scenario, we demonstrate the application of SARP to a mobile robot navigation task in a hospital simulation, see Fig. \mbox{\ref{fig:hospital}}. 
Our repair method guides the robot towards safe behavior by incorporating safety constraints on its predicted actions and the outputs of a trained predictive model that predicts possible collisions using range sensor readings. 
By doing so, we mitigate the risk of accidents or collisions, which is crucial for robot navigation in a safety-critical environment such as a hospital.}
}{
The first example is presented as a simulation showcase, comparing SARP with two state-of-the-art safe RL approaches: Constrained Policy Optimization (CPO) \cite{achiam2017constrained} and the safety-layer approach (DDPG+SL) \cite{dalal2018safe}.
In this scenario, we demonstrate the application of SARP to a mobile robot navigation task in a hospital simulation, see Fig. \ref{fig:hospital}. 
Our repair method guides the robot towards safe behavior by incorporating safety constraints on its predicted actions and the outputs of a trained predictive model that predicts possible collisions using range sensor readings. 
By doing so, we mitigate the risk of accidents or collisions, which is crucial for robot navigation in a safety-critical environment such as a hospital.}

The second scenario demonstrates a real-world application of SARP in controlling a lower-leg prosthesis device.
Recently, machine learning has been shown to be a promising approach in enhancing the control and functionality of lower-leg prosthetic devices \cite{gehlhar2023review}.
Here we aim to limit the action rate and the pressure applied to the foot, see Fig. \ref{fig:sensor model} (right). 
\ifthenelse{\boolean{extend}}{Excessive forces applied to the heel during walking can lead to joint pain, stiffness in knees and hips, as well as an increased risk of musculoskeletal disorders or osteoarthritis \cite{liikavainio2007loading}.}{}
Measuring foot forces in real-time can be expensive due to several factors, such as the need for multiple sensors for accurate measurements as well as the complexity of integration with the prosthesis and customized software.
Hence, we utilize a predictive model to estimate the pressures applied to the foot. 
We demonstrate that SARP promotes the safe operation of the prosthesis device by guiding it to behave within safe boundaries through enforcing safety constraints on the predicted actions and bio-mechanical states of the system.
\ifthenelse{\boolean{extend}}{In all experiments, we defined {$\mathcal{L} = ||\pi_{\theta} - \pi_{\theta_{IL}}||^2$}.}{
\ifthenelse{\boolean{highlight}}{
\hlcyan{In all experiments, we defined $\mathcal{L} = ||\pi_{\theta} - \pi_{\theta_{IL}}||^2$.}}{In all experiments, we defined {$\mathcal{L} = ||\pi_{\theta} - \pi_{\theta_{IL}}||^2$}.}}
\ifthenelse{\boolean{highlight}}{\hl{Throughout our experiments, we calculate all relative improvement metrics, including Relative Change (RC) and Side Effect (SE), using the following formula (assuming the target value is $x$):
$(x_{new} - x_{orig})/(x_{orig})\times 100\%$.}}{Throughout our experiments, we calculate all relative improvement metrics, including Relative Change (RC) and Side Effect (SE), using the following formula (assuming the target value is $x$):
$(x_{new} - x_{orig})/(x_{orig})\times 100\%$.
}
\ifthenelse{\boolean{highlight}}{
\begin{table}[tb]
\vspace{7pt}
\caption{\hl{Robot navigation statistics. GR: average percentage of the trajectories that reached the goal, E: average percentage of safe samples (Efficacy), and ST: simulation time. The results are the average of 200 test trajectories. } }
\label{tab:navigation}

\begin{adjustbox}{width=0.45\textwidth}
\setlength{\aboverulesep}{0pt}
\setlength{\belowrulesep}{1pt}
\setlength{\extrarowheight}{.7ex}
\begin{tabular}{
lccc
}
\toprule
    Method &GR (Collision) & E (Collision) & ST [$h$] \\ \midrule
Orig. Policy 
        & $68.7\%$& $74.4\%$& $0${\tiny$+0.5$} 
        \\ \hline\hline
SARP$_{\text{penalty}}$
        & $89.8\%$ & $92.7\%$& \cellcolor[HTML]{EFEFEF}$\mathbf{3}${\tiny$+0.5$} 
        \\
SARP$_{\text{Lagrangian}}$
        & \cellcolor[HTML]{EFEFEF}$\mathbf{98.1\%}$ & $97.8\%$&  \cellcolor[HTML]{EFEFEF}$\mathbf{3}${\tiny$+0.5$} 
        \\
CPO$_{\text{sparse, 3h}}$
        & $74.4\%$ & $99.3\%$&  \cellcolor[HTML]{EFEFEF}$\mathbf{3}${\tiny$+0.5$} 
        \\
CPO$_{\text{dense, 3h}}$
        & $75.0\%$ & $97.2\%$ &  \cellcolor[HTML]{EFEFEF}$\mathbf{3}${\tiny$+0.5$} 
        \\ 
DDPG+SL$_{\text{sparse, 3h}}$
        & $78.2\%$ & $99.8\%$ &  $\mathbf{{\color{red} 3}}\!+\!3${\tiny$+0.5$} 
        \\
DDPG+SL$_{\text{dense, 3h}}$
        & $72.1\%$ & $99.7\%$ &  $\mathbf{{\color{red} 3}}\!+\!3${\tiny$+0.5$}  
        \\
\hline\hline
CPO$_{\text{sparse}, 98\%}$
        & $96.8\%$ & $99.7\%$ & {\color{red}$\mathbf{25}$}{\tiny$+0.5$} 
        \\
CPO$_{\text{dense}, 98\%}$
        & $92.6\%$ & $99.4\%$ & {\color{red}$\mathbf{18}$}{\tiny$+0.5$} 
        \\
DDPG+SL$_{\text{sparse}, 98\%}$
        & $94.1\%$ & \cellcolor[HTML]{EFEFEF}$\mathbf{99.9\%}$ & $\mathbf{{\color{red}21}\!+\!{\color{black} 3}}${\tiny$+0.5$} 
        \\
DDPG+SL$_{\text{dense}, 98\%}$
        & $90.4\%$ & \cellcolor[HTML]{EFEFEF}$\mathbf{99.9\%}$ &  $\mathbf{{\color{red}30}\!+\!{\color{black} 3}}${\tiny$+0.5$}  
        \\
        \bottomrule 
\end{tabular}%

\end{adjustbox}
\vspace{-5pt}
\end{table}
}
{
\begin{table}[tb]
\vspace{7pt}
\caption{Robot navigation statistics. GR: average percentage of the trajectories that reached the goal, E: average percentage of safe samples (Efficacy), and ST: simulation time. The results are the average of 200 test trajectories. }
\label{tab:navigation}

\begin{adjustbox}{width=0.45\textwidth}
\setlength{\aboverulesep}{0pt}
\setlength{\belowrulesep}{1pt}
\setlength{\extrarowheight}{.7ex}
\begin{tabular}{
lccc
}
\toprule
    Method &GR (Collision) & E (Collision) & ST [$h$] \\ \midrule
Orig. Policy 
        & $68.7\%$& $74.4\%$& $0${\tiny$+0.5$} 
        \\ \hline\hline
SARP$_{\text{penalty}}$
        & $89.8\%$ & $92.7\%$& \cellcolor[HTML]{EFEFEF}$\mathbf{3}${\tiny$+0.5$} 
        \\
SARP$_{\text{Lagrangian}}$
        & \cellcolor[HTML]{EFEFEF}$\mathbf{98.1\%}$ & $97.8\%$&  \cellcolor[HTML]{EFEFEF}$\mathbf{3}${\tiny$+0.5$} 
        \\
CPO$_{\text{sparse, 3h}}$
        & $74.4\%$ & $99.3\%$&  \cellcolor[HTML]{EFEFEF}$\mathbf{3}${\tiny$+0.5$} 
        \\
CPO$_{\text{dense, 3h}}$
        & $75.0\%$ & $97.2\%$ &  \cellcolor[HTML]{EFEFEF}$\mathbf{3}${\tiny$+0.5$} 
        \\ 
DDPG+SL$_{\text{sparse, 3h}}$
        & $78.2\%$ & $99.8\%$ &  $\mathbf{{\color{red} 3}}\!+\!3${\tiny$+0.5$} 
        \\
DDPG+SL$_{\text{dense, 3h}}$
        & $72.1\%$ & $99.7\%$ &  $\mathbf{{\color{red} 3}}\!+\!3${\tiny$+0.5$}  
        \\
\hline\hline
CPO$_{\text{sparse}, 98\%}$
        & $96.8\%$ & $99.7\%$ & {\color{red}$\mathbf{25}$}{\tiny$+0.5$} 
        \\
CPO$_{\text{dense}, 98\%}$
        & $92.6\%$ & $99.4\%$ & {\color{red}$\mathbf{18}$}{\tiny$+0.5$} 
        \\
DDPG+SL$_{\text{sparse}, 98\%}$
        & $94.1\%$ & \cellcolor[HTML]{EFEFEF}$\mathbf{99.9\%}$ & $\mathbf{{\color{red}21}\!+\!{\color{black} 3}}${\tiny$+0.5$} 
        \\
DDPG+SL$_{\text{dense}, 98\%}$
        & $90.4\%$ & \cellcolor[HTML]{EFEFEF}$\mathbf{99.9\%}$ &  $\mathbf{{\color{red}30}\!+\!{\color{black} 3}}${\tiny$+0.5$}  
        \\
        \bottomrule 
\end{tabular}%

\end{adjustbox}
\vspace{-5pt}
\end{table}
}
\subsection{Showcase Example: Robot Navigation in Hospital}
\label{subsec: experimental setup}
In this task, a NN policy is trained to control the linear and angular velocities of a robot, denoted as $a=[v(t),\omega(t)]^T$. 
The system states include the robot's Cartesian goal location $[x_g,y_g]^T$, Euclidean distance to the goal $d_g$, heading towards the goal $\phi_g$,
and a 2D range sensor readings vector $r=[r_1,\cdots,r_M]^T$, with $M$ representing the number of sensor rays ($M=10$ in our experiments). 
Hence, the system state is defined as $s=[x_g(t),y_g(t),d_g(t), \phi_g(t), r(t)]^T$. 
To predict collisions, we utilize a predictive model $f(r(t))$ that takes the range sensor readings at time $t$ and predicts a collision occurrence at $t+1$, 
i.e., $\feat=\{0,1\}$, where $1$ indicates a collision and $0$ indicates no collision.
In this experiment, we utilized the TurtleBot3 robot model \cite{turtlebot3} within a Gazebo simulated Hospital World\footnote{We employed the Gazebo simulated Hospital World, a publicly available environment developed by Amazon Web Services, accessible at \url{https://github.com/aws-robotics/aws-robomaker-small-warehouse-world.git}}, as shown in Fig. \ref{fig:hospital}.

For training the original policy $\pi_{\theta}$, we first manually moved the robot using a joystick from two initial positions towards six designated goals, as depicted in Fig. \ref{fig:hospital_nav}. 
We gathered a total of $5$ trajectories for each combination of initial and goal positions. 
\ifthenelse{\boolean{highlight}}{
\hlcyan{We allocated a fixed duration of $0.5$ hours for collecting expert demonstrations.}}{
we allocated a fixed duration of $0.5$ hours for collecting expert demonstrations.}
The collected trajectories are then employed to train a two-hidden-layer policy with 256 ReLU nodes at each hidden layer using Behavioral Cloning. 
The collision predictive model $f(r(t))$ is assumed to be a one-hidden-layer neural network with 128 ReLU hidden nodes and a softmax final layer.
For training the collision model, we allowed the robot to move randomly in the environment for a duration of $3$ hours using the wander steering algorithm \cite{reynolds1999steering}.
We finally repaired the policy using Algorithm \ref{alg:verifier2} over the collected expert trajectories to ensure that $\feat=0$ for all states in the training set.
During the repair process, the initial values were set as $\mu_0=5$, $\eta=0.001$, and $\beta=1.5$.
In experiments, we optimized the policy using the full loss term from Eq. (\ref{eq: augmented loss}) as SARP$_{\text{Lagrangian}}$, and only the quadratic penalty term ($\lambda=0$) as SARP$_{\text{penalty}}$.
Figure {\ref{fig:hospital_nav}} shows the robot trajectories before (a) and after (b) repair.

\ifthenelse{\boolean{highlight}}
{\hl{We conducted a performance comparison between SARP and two safe RL methods of
CPO\mbox{\cite{achiam2017constrained}}, and DDPG+SL\mbox{\cite{dalal2018safe}}. 
CPO considers the constraints in policy optimization and employs conditional gradient descent with line search to perform policy updates.
DDPG+SL optimizes the policy with the Deep Deterministic Policy Gradient (DDPG) method and incorporates a filter to direct actions toward safety during policy exploration.}}
{We conducted a performance comparison between SARP and two safe RL methods of CPO \cite{achiam2017constrained}, 
and DDPG+SL \cite{dalal2018safe}. 
CPO considers the constraints in policy optimization and employs conditional gradient descent with line search to perform policy updates.
DDPG+SL optimizes the policy with the Deep Deterministic Policy Gradient (DDPG) method and incorporates a filter to direct actions toward safety during policy exploration.}
For a fair comparison, we initialized the RL policies with the original policy that we trained using expert demonstrations.
We also used the samples that we collected for training the collision model to pre-train the safety filter in DDPG+SL.
\ifthenelse{\boolean{highlight}}{
\hlcyan{In this study, we define the simulation time (ST) as the duration that the robot interacts with the environment to collect samples for policy optimization. 
Therefore, in total, SARP utilized $0.5~h$ for collecting expert demonstrations and $3~h$ for samples collection used in training the collision model.
Since DDPG+SL's policy and safety filter are initialized with the same samples, it used $3+0.5~h$ of simulation time for initialization. 
In contrast, CPO only required $0.5~h$ for initialization, as it evaluates collisions during the training process. 
The remaining simulation time for DDPG+SL and CPO is dedicated to their training process.}}{
In this study, we define the simulation time (ST) as the duration that the robot interacts with the environment to collect samples for policy optimization. 
Therefore, in total, SARP utilized $0.5~h$ for collecting expert demonstrations and $3~h$ for samples collection used in training the collision model.
Since DDPG+SL's policy and safety filter are initialized with the same samples, it used $3+0.5~h$ of simulation time for initialization. 
In contrast, CPO only required $0.5~h$ for initialization, as it evaluates collisions during the training process. 
The remaining simulation time for DDPG+SL and CPO is dedicated to their training process.}
In CPO and DDPG+SL, we used two reward functions: sparse and dense. The sparse reward assigns the highest value when the robot reaches the goal and zero otherwise. The dense reward function rewards the robot for reaching the goal and penalizes it based on the change in distance to the goal, $d_g(t-1)-d_g(t)$.
For comparison, we tested the optimized policies on a larger number of trajectories ($200$ trajectories).
\ifthenelse{\boolean{highlight}}{
\hlcyan{To assess policy performance, we defined two metrics: 
\textit{Goal-reaching (GR)}, measuring the percentage of trajectories reaching goals, and 
\textit{Efficacy (E)}, evaluating sample-wise collision satisfaction by considering samples with range readings below $0.3$ meters as collisions.}}{
To assess policy performance, we defined two metrics: 
\textit{Goal-reaching (GR)}, measuring the percentage of trajectories reaching goals, and 
\textit{Efficacy (E)}, evaluating sample-wise constraint satisfaction by considering samples with range readings below $0.3$ meters as collisions.}

\noindent \textbf{Evaluation.}\label{sec:eval-hospital}
\ifthenelse{\boolean{highlight}}
{\hlcyan{We first evaluated the accuracy of CPO, and DDPG+SL by training them for $3~h$ (similar simulation time SARP used for collecting collision trajectories). 
We terminate the RL episode 10 epochs following a collision.
The results presented in Table \mbox{\ref{tab:navigation}} indicate that SARP outperforms the safe RL methods in terms of GR, showing an improvement of approximately $43\%$ compared to $14\%$ in RL methods.
However, the RL methods demonstrate slightly better performance in terms of E, with a $2\%$ improvement compared to SARP. 
We further extended the evaluation of RL methods until they reached a GR of $98\%$ in training, which was the highest achieved by SARP as highlighted in Table \ref{tab:navigation}. 
In terms of GR, SARP outperforms CPO and DDPG+SL by $1.3\%$.
CPO and DDPG+SL require at least an additional 15 hours compared to SARP to achieve this level of GR accuracy.}}
{We first evaluated the accuracy of CPO, and DDPG+SL by training them for $3~h$ (similar simulation time SARP used for collecting collision trajectories). 
We terminate the RL episode 10 epochs following a collision.
The results presented in Table \mbox{\ref{tab:navigation}} indicate that SARP outperforms the safe RL methods in terms of GR, showing an improvement of approximately $43\%$ compared to $14\%$ in RL methods.
However, the RL methods demonstrate slightly better performance in terms of E, with a $2\%$ improvement compared to SARP. 
We further extended the evaluation of RL methods until they reached a GR of $98\%$ in training, which was the highest achieved by SARP as highlighted in Table \ref{tab:navigation}. 
In terms of GR, SARP outperforms CPO and DDPG+SL by $1.3\%$.
CPO and DDPG+SL require at least an additional 15 hours compared to SARP to achieve this level of GR accuracy.}
\ifthenelse{\boolean{highlight}}{
\begin{figure}[tb]
\vspace{7pt}
    \centering
    \includegraphics[width=.43\textwidth]{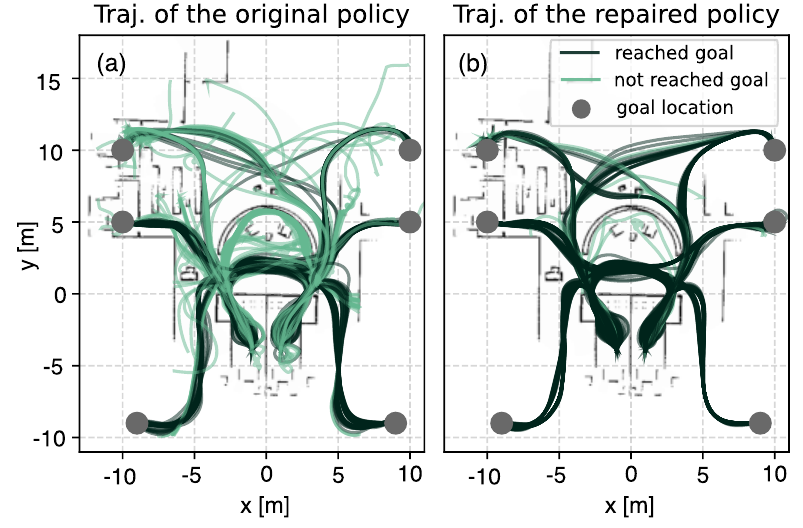}
    \caption{\hl{Navigation in hospital: the collision avoidance of the robot improves by over $93\%$ with SARP shown in (b) compared to the original policy depicted in (a). 
    }
    }
    \label{fig:hospital_nav}
    \vspace{-10pt}
\end{figure}
}{
\begin{figure}[tb]
\vspace{7pt}
    \centering
    \includegraphics[width=.43\textwidth]{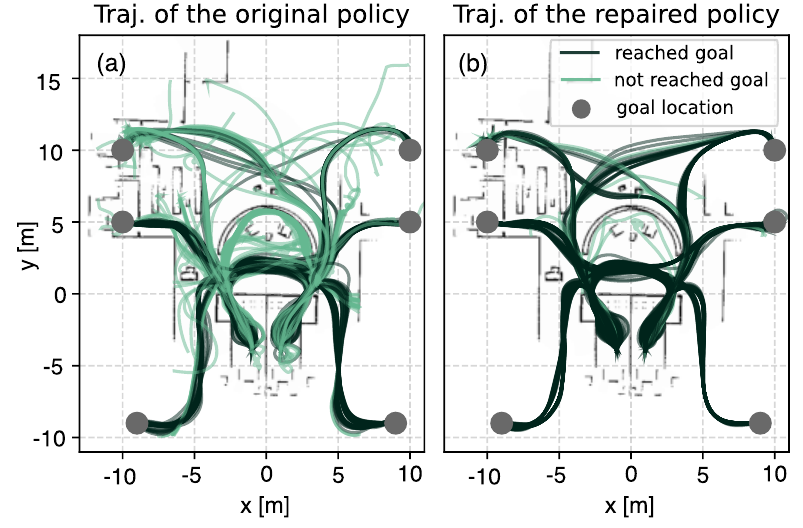}
    \caption{Navigation in hospital: the collision avoidance of the robot improves by over $93\%$ with SARP shown in (b) compared to the original policy depicted in (a).
    }
    \label{fig:hospital_nav}
    \vspace{-14pt}
\end{figure}
}
\ifthenelse{\boolean{highlight}}{
\hl{
In Fig. {\ref{fig:hospital_nav2}} (a)-(b), we demonstrate our framework's ability to enforce multiple constraints, preventing collisions and keeping velocities below $0.9~[m/s]$.
After repair, minimum scan values consistently exceeded $0.3$, and velocities stayed below $0.9~[m/s]$. 
We also investigated the relationship between GR and E versus the collision model's accuracy in Fig. {\ref{fig:hospital_nav2}} (c). 
To generate Fig. {\ref{fig:hospital_nav2}} (c), we conducted training on 5 networks for each level of collision accuracy. 
Moreover, we tested these networks with 200 trajectories each.
The results demonstrate that the percentage of safe samples remains above $94\%$ across various collision model accuracies. 
However, GR drops below $90\%$ when the collision model is less than $96\%$ accurate.
While SARP improves constraint satisfaction over the original policy ($74.4\%$), employing an inaccurate collision model leads to a deviation from the original task.
Thus, SARP's accuracy can be influenced by the accuracy of the predictive model. 
}
}{
In Fig. {\ref{fig:hospital_nav2}} (a)-(b), we demonstrate our framework's ability to enforce multiple constraints, preventing collisions and keeping velocities below $0.9~[m/s]$.
After repair, minimum scan values consistently exceeded $0.3$, and velocities stayed below $0.9~[m/s]$. 
We also investigated the relationship between GR and E versus the collision model's accuracy in Fig. {\ref{fig:hospital_nav2}} (c). 
To generate Fig. {\ref{fig:hospital_nav2}} (c), we conducted training on 5 networks for each level of collision accuracy. 
Moreover, we tested these networks with 200 trajectories each.
The results demonstrate that the percentage of safe samples remains above $94\%$ across various collision model accuracies. 
However, GR drops below $90\%$ when the collision model is less than $96\%$ accurate.
While SARP improves constraint satisfaction over the original policy ($74.4\%$), employing an inaccurate collision model leads to a deviation from the original task.
Thus, SARP's accuracy can be influenced by the accuracy of the predictive model.}
\ifthenelse{\boolean{highlight}}{
\begin{figure}[tb]
\vspace{7pt}
    \centering
    \includegraphics[width=.41\textwidth]{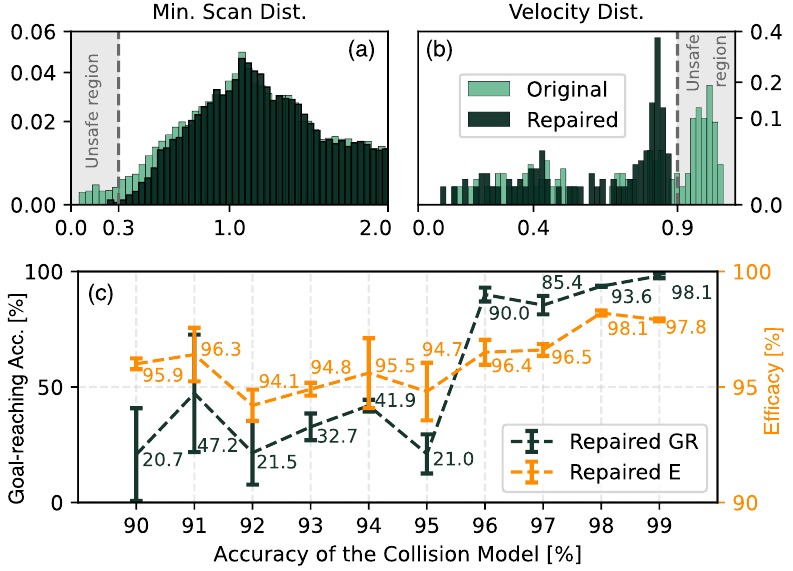}
    \caption{\hl{
   Navigation in hospital: (a)-(b) minimum range sensor values and velocity distributions before and after constraint application. (c) Goal-reaching accuracy and the percentage of safe samples vs. the accuracy of the prediction model.
    }}
    \label{fig:hospital_nav2}
    \vspace{-5pt}
\end{figure}
}{
\begin{figure}[tb]
\vspace{7pt}
    \centering
    \includegraphics[width=.41\textwidth]{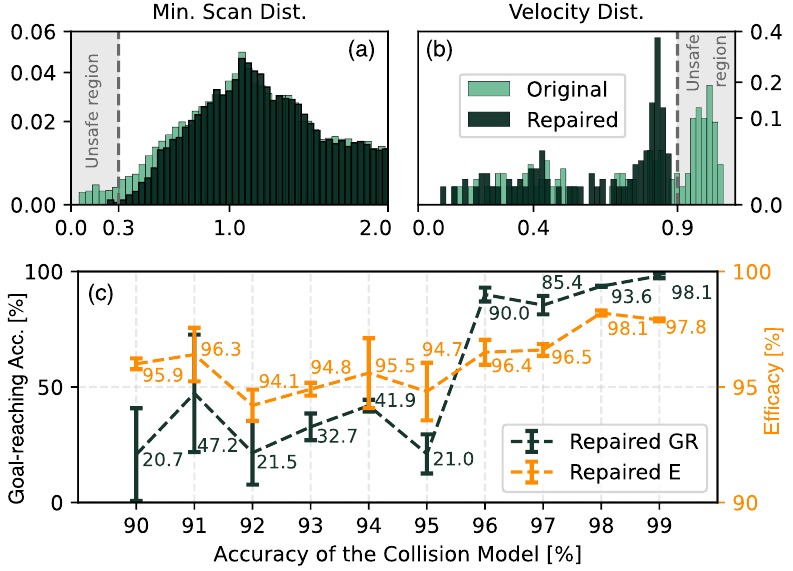}
    \caption{Navigation in hospital: (a)-(b) minimum range sensor values and velocity distributions before and after constraint application. (c) Goal-reaching accuracy and the percentage of safe samples vs. the accuracy of the prediction model.}
    \label{fig:hospital_nav2}
    \vspace{-5pt}
\end{figure}
}
\begin{figure}[tb]
    \vspace{7pt}
    \centering
    \includegraphics[scale=.49]{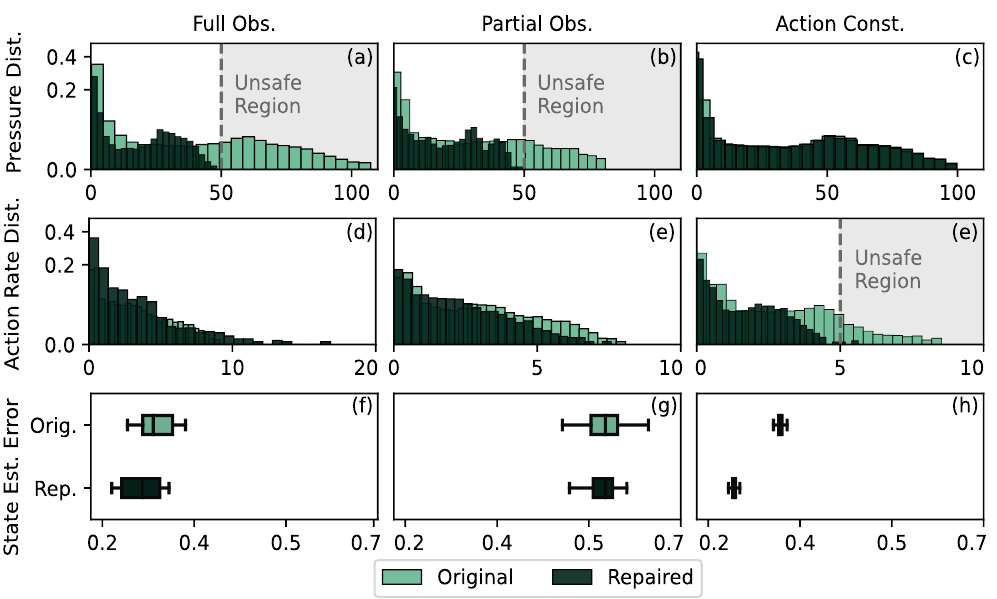}
    \caption{Real-world walking results: 
    (Top) distribution of pressure values, 
    (Middle) distribution of action rate values,
    (Bottom) state estimation error.}
    \label{fig:results}
    \vspace{-12pt}
\end{figure}
\subsection{Real-world Example: Lower-leg Prosthesis Control}
In this task, policy $\pi_{\theta}$ takes as input a history of the previous $h$ observable state values at time step $t$, i.e., $s = [s^o(t-h+1),\cdots,s^o(t-1), s^o(t)]$. The policy then predicts the future ankle angles $a=\left[\alpha_a(t), \alpha_a(t+1), \cdots, \alpha_a(t+q-1)\right]$ of the prosthesis device for $q$ steps ahead.
To predict the system's states for the $q$ steps ahead, we use the predictive model $f(s,a)$. 
Specifically, $\feat = \left[s^f(t+1), s^f(t+2), \cdots, s^f(t+q+1)\right]$, where the states are the angle and velocity of the upper and lower limb, and the pressure sensor insole readings, i.e., $s^f=[\alpha_{ul},\Dot{\alpha}_{ul}, \alpha_{ll},\Dot{\alpha}_{ll}, p]^T$, respectively. The states and actions of the system are shown in Fig.~\ref{fig:sensor model} (left) and (middle).
In our experiments, we only considered the sensors located on the heel as part of system states, $p=[p_1,\cdots,p_4]^T$.

We investigate two modes of operation: fully observable and partially observable. 
In the fully observable mode, the policy predicts future actions given complete sensor readings, represented as $s^o=s^f$. 
Conversely, in the partially observable mode, the policy receives upper and lower limb sensor readings only, denoted as $s^o=[\alpha_{ul},\Dot{\alpha}_{ul}, \alpha_{ll},\Dot{\alpha}_{ll}]^T$, while assuming the absence of pressure readings during testing.

In this experiment, we evaluate our approach on three test cases: 
bounding pressure for both fully observable and partially observable scenarios
, and bounding the action rate solely for the fully observable case. 
For bounding pressure, we ensure that the pressure readings remain below $50~[N/cm^2]$. This constraint can be defined as $\mathcal{G}_p = \{p_j(t+i)<50\text{ for $j=1,\cdots,4$}\}_{i=0}^q$.
Regarding the constraint on the action rate, we aim to limit the rate of change in ankle angle to prevent abrupt actions and promote smoother transitions. 
We enforce a condition that restricts the absolute difference between consecutive ankle angles to be below $5~[deg/s]$, denoted as $\mathcal{G}_{\alpha_a}=\{\lvert \alpha_a(t+i+1) - \alpha_a(t+i)\rvert \leq 5\}^q_{i=0}$.
In all experiments, we specified $q=30$ and $h=10$.

In training, we first collected data by recording the walking gait of a healthy individual who was not using a prosthetic device under a study approved by the Institutional Review Board (IRB). 
We used inertial measurement units (IMUs) to capture the joint angles and pressure sensor insole to measure foot pressures.
We pre-trained a two-hidden-layer policy and a one-hidden-layer predictive model with 512 ReLU nodes at each hidden layer using BC and supervised learning, respectively.
We repaired the policy until all repair samples satisfied the constraints following the Algorithm \ref{alg:verifier2}. 
The first predicted ankle angle $\alpha_a(t)$ 
\ifthenelse{\boolean{extend}}{(receding horizon strategy)}{}
was then used as the control parameter for a PD controller on the prosthetic device. 
 We specified $\mu_0=5$, $\beta=5$ and $\eta = 0.001$. 
\ifthenelse{\boolean{extend}}{We tested the repaired policies for 10 minutes of walking.
We used an ankle bypass to allow the able-bodied subject to walk on the prosthesis and to generate strides outside the original training distribution \cite{cortino2022stair}. }{}

\begin{table}[tb]
\vspace{7pt}
\caption{Policy repair statistics offline. This table reports RC: average Relative Change of state or action values with respect to their original values after repair on test data, E: average percentage of unsafe samples that are repaired (Efficacy), and SE: average relative change of prediction error after repair with respect to the original error evaluated over the originally safe state and action regions (Side Effect). }
\label{tab:my-table}

\begin{adjustbox}{width=0.45\textwidth}
\setlength{\aboverulesep}{0pt}
\setlength{\belowrulesep}{1pt}
\setlength{\extrarowheight}{.75ex}

\begin{tabular}{@{}
cccccc
@{}}
\toprule
    &\makecell{RC \\(Pressure)} & \makecell{RC \\(Action)} & E  &  \makecell{SE \\(State)} & \makecell{SE \\(Action)}  \\ \midrule
Full Obs. 
        & \cellcolor[HTML]{EFEFEF}$\mathbf{425\%}$& $11\%$& \cellcolor[HTML]{EFEFEF}$\mathbf{99\%}$&  \cellcolor[HTML]{EFEFEF}$\mathbf{-40\%}$ & $2\%$
        \\
Partial Obs.
        & \cellcolor[HTML]{EFEFEF} $\mathbf{110\%}$ & $4.5\%$& \cellcolor[HTML]{EFEFEF}$\mathbf{100\%}$ &  \cellcolor[HTML]{EFEFEF}$\mathbf{-23\%}$ & $30\%$
        \\
Action Const.  
        & $13\%$ & \cellcolor[HTML]{EFEFEF} $\mathbf{118\%}$ & \cellcolor[HTML]{EFEFEF}$\mathbf{99\%}$ &  \cellcolor[HTML]{EFEFEF}$\mathbf{-5}\%$ & $7\%$
        \\ 
        \bottomrule 
\end{tabular}%
\end{adjustbox}
\vspace{-5pt}
\end{table}

\noindent \textbf{Evaluataion.}
We evaluated the performance of our method in two online and offline cases. 
Table \ref{tab:my-table} shows the offline evaluation of repaired versus originally trained models.
We assessed the models using the test samples.
In the fully observable (Full Obs.) and partially observable (Partial Obs.) test cases, the pressure values are changed by over $100\%$ compared to the original model.
This is because the repaired policy prevents the pressure values from exceeding a certain threshold.
Bounding the action rate also changed the action values by $118\%$, as illustrated in Table \ref{tab:my-table}.
The repair demonstrates over $99\%$ efficacy in all models, indicating that constraint satisfaction is well-generalized to the testing samples.
\ifthenelse{\boolean{highlight}}{\hl{Finally, we evaluated the relative change in the prediction error of predictive
and policy networks for the input regions that were originally safe (SE). 
We assessed this metric over the original testing samples.
This metric is important as it measures if the repaired policy has side effects on the prediction
performance of the regions that do not need to be repaired.}}{Finally, we evaluated the relative change in the prediction error of predictive
and policy networks for the input regions that were originally safe (SE). 
We assessed this metric over the original testing samples.
This metric is important as it measures if the repaired policy has side effects on the prediction
performance of the regions that do not need to be repaired.}
Our results indicate that repair improved the state prediction error by up to $40\%$ in the safe regions. 
However, the action error is increased in all three cases.
This could be attributed to predicting actions for a future time horizon, that considers manipulating actions in advance of unsafe behavior occurrence.

\begin{figure}[t]
    \vspace{7pt}
    \centering
    \includegraphics[width=0.93\linewidth]{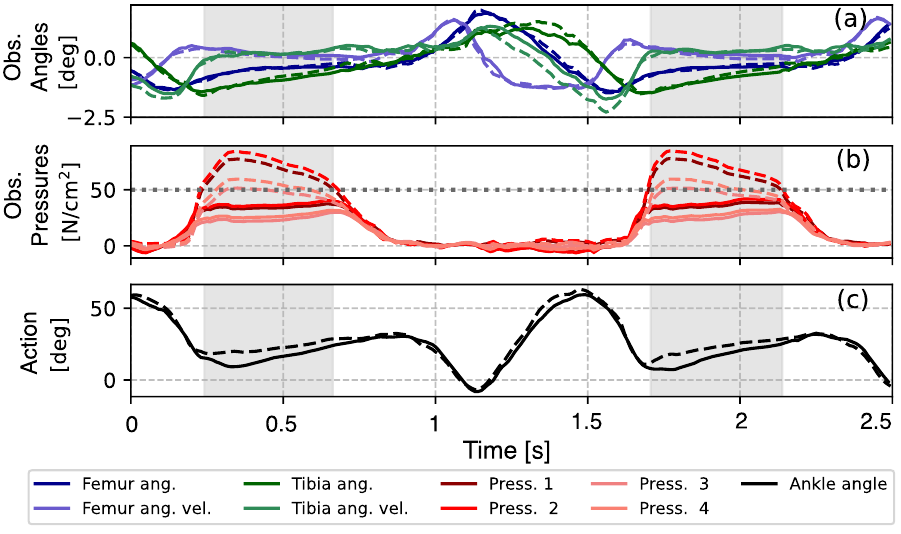}
    \caption{Online walking signals: \textbf{fully observable case with bounded pressure values $p\leq 50 ~[N/cm^2]$}. Dashed lines show the walking signals with the original model and the solid lines show the repaired signals.}
    \label{fig:signal obs}
    \vspace{-7pt}
\end{figure}
\begin{figure}[t]
    \vspace{7pt}
    \centering
    \includegraphics[width=0.95\linewidth]{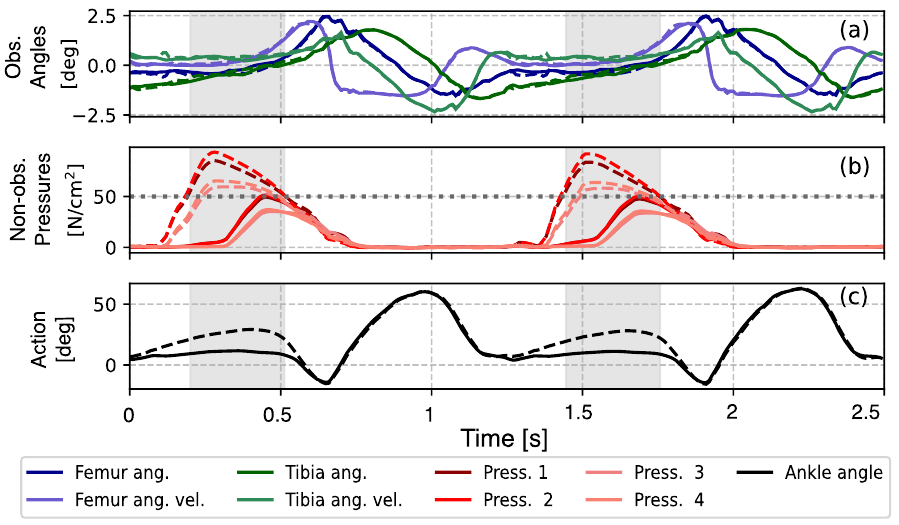}
    \caption{Online walking signals: \textbf{partially observable case with bounded pressure values $p\leq 50 ~[N/cm^2]$}. Dashed lines show the walking signals with the original model and the solid lines show the repaired signals.}
    \label{fig:signalNonobs}
    \vspace{-11pt}
\end{figure}
\begin{figure}[t]
    \vspace{7pt}
    \centering
    \includegraphics[width=0.94\linewidth]{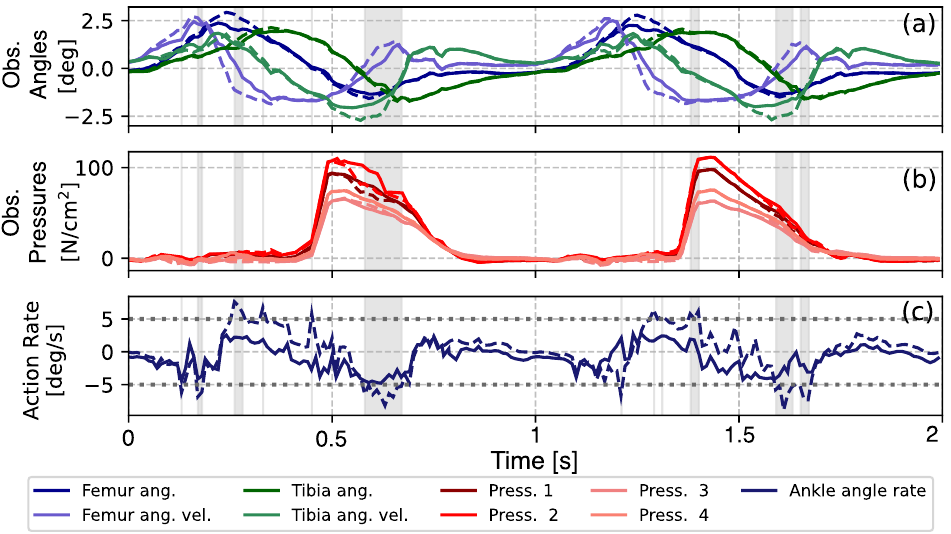}
    \caption{Online walking signals: \textbf{fully observable case with bounded action rate $|\Delta \alpha_a|\leq 5 ~[deg/s]$}. Dashed lines show the walking signals with the original model and the solid lines show the repaired signals.}
    \label{fig:signalCtrl}
    \vspace{-11pt}
\end{figure}
Figure \ref{fig:results} shows the online walking results. 
The pressure distributions demonstrate that the pressure constraints are successfully satisfied for the fully observable and partially observable cases with bounded pressure values.
The pressure values are bounded by $50~ [N/cm^2]$ as shown in Figures \ref{fig:results} (a)-(b).
The real-world walking results of the action rate constrained model also show that the action rates never exceed $5~[deg/s]$, see Fig. \ref{fig:results} (e).
The state estimation error average is below $0.35$ in fully observable and action-bounded test cases, Figures \ref{fig:results} (f) and (h), respectively. 
Nonetheless, the mean state estimation error exhibits a nearly 0.3 increase in the partially observable case, as illustrated in Fig. \ref{fig:results} (g).
One possible explanation for this outcome is the inability of the network to capture the temporal influence of non-observable states during training as inputs. Despite this, the state estimation accuracy remains reasonably high, as evidenced by the successful pressure bounding at the heel achieved by the policy repair.
Figures \ref{fig:signal obs}, \ref{fig:signalNonobs}, and \ref{fig:signalCtrl} show the online walking signals for the fully observable case with bounded pressure, the partially observable test with bounded pressure, and the bounded action rate case, respectively.
The pressure values are effectively constrained in both completely observable and partially observable situations, as depicted in Figures \ref{fig:signal obs} and \ref{fig:signalNonobs}. 
In Fig. \ref{fig:signalCtrl}, the repaired model bounds the action rate to $5~[\deg/s]$ with success.


\subsection{Discussion}
\label{sec:discussion}
Our experiments in Sec. \ref{sec:eval-hospital} showed that the accuracy of predictive model might impact the safety accuracy of the repaired policy.
\ifthenelse{\boolean{highlight}}{\hlcyan{We also recognize that RL enables agents to explore and uncover novel solutions that may not be evident in expert demonstrations or predefined safety constraints.
However, repairing policies with predictive models offers several benefits that position SARP as a reliable approach for continuous policy monitoring and repair.}}{We also recognize that RL enables agents to explore and uncover novel solutions that may not be evident in expert demonstrations or predefined safety constraints.
However, repairing policies with predictive models offers several benefits that position SARP as a reliable approach for continuous policy monitoring and repair.}
\ifthenelse{\boolean{extend}}{
Firstly, it eliminates the need to explicitly estimate or learn the system's transition model, e.g., $\sigma(s(t),a(t))=s(t+1)$.
Instead, we can focus on immediately predicting the variables of interest. In our case, these variables include the physical interaction between the robot and colliders in the environment, and the bio-mechanical features.
Secondly, it allows monitoring of system safety using low-dimensional feature indicators, eliminating the need to track high-dimensional states to optimize policy safety violations across trajectories. 
Furthermore, predictive models enable proactive identification and mitigation of potential safety issues, promoting a proactive safety approach. 
It allows reasoning about the safety of after-effects and ramifications of impacts, such as assessing whether the current action would result in a collision.
Finally, predictive models enhance learning efficiency by reducing the need for time-consuming simulations. Predicting system behavior and informed decisions saves time and resources in training safe and effective robot policies.}{
Firstly, it eliminates the need to explicitly estimate or learn the system's transition model, e.g., $\sigma(s(t),a(t))=s(t+1)$.
Instead, we can focus on immediately predicting the (lower-dimensional) features of interest. In our case, these variables include the physical interaction between the robot and colliders in the environment, and the bio-mechanical features.
Furthermore, predictive models enable proactive identification and mitigation of potential safety issues, promoting a proactive safety approach. 
It allows reasoning about the safety of after-effects and ramifications of impacts, such as assessing whether the current action would result in a collision.
\ifthenelse{\boolean{highlight}}{\hlcyan{
Finally, predictive models enhance learning efficiency by reducing the need for time-consuming simulations. 
This positions SARP as a suitable method to address safety concerns in scenarios where extensive exploration of safety policies is not feasible.}}{
Finally, predictive models enhance learning efficiency by reducing the need for time-consuming simulations. 
This positions SARP as a suitable method to address safety concerns in scenarios where extensive exploration of safety policies is not feasible.}}

Controlling lower-leg prostheses serves as a compelling application for SARP in robot learning, addressing challenging dynamics of human-robot interactions and diverse safety concerns.
In our current IRB-approved study, we focused on a single human subject. 
However, further research is required to evaluate and analyze the performance of SARP-repaired policies across multiple subjects.
Moreover, including additional biomechanical indices,
such as ground reaction forces or joint forces, can enhance policy safety and robustness under SARP.
\ifthenelse{\boolean{extend}}{Future studies will explore these aspects.}{}
\ifthenelse{\boolean{highlight}}{\hl{Finally, we tested the policy repair process for the prosthesis example with the same hyperparameters as in the hospital scenario. We observed similar performance outcomes after 200 epochs. This suggests that repair performance is not sensitive to parameter variations. In future work, we will analyze repair sensitivity to optimization parameter selection and compare it with other algorithms.}}
{Finally, we tested the policy repair process for the prosthesis example with the same hyperparameters as in the hospital scenario. 
We observed similar efficacy outcomes after 200 epochs. 
In future work, we will further analyze repair sensitivity to optimization parameter selection and compare it with other algorithms.}

\section{Conclusions and Future Work}
\label{sec:conclusion}
Our paper presents a method that combines imitation learning with neural network repair to address safety concerns in robot learning. 
By leveraging predictive models, we repair policies to adhere to predefined safety constraints, enhancing safety and adaptability in robot behavior. 
We showcase the effectiveness of our approach in robot collision avoidance and lower-leg prosthesis control scenarios.
Future work includes exploring continual learning methods to update the predictive model based on new experiences and adapt the repaired policy to changing environmental conditions and evolving safety requirements. 



\bibliographystyle{IEEEtran}
\bibliography{root.bib}

\end{document}